\pdfoutput=1

\documentclass[11pt]{article}

\usepackage[final]{acl}

\usepackage{times}
\usepackage{latexsym}

\usepackage[T1]{fontenc}

\usepackage[utf8]{inputenc}

\usepackage{microtype}

\usepackage{inconsolata}

\usepackage{graphicx}
\usepackage{subcaption}
\usepackage{graphicx}

\usepackage{booktabs}
\usepackage{multirow}
\usepackage{makecell}
\usepackage{caption}
\usepackage{amssymb}
\usepackage{lipsum,afterpage,refcount}
\usepackage{tikz,lipsum}
\usepackage[most]{tcolorbox}
\usepackage{subcaption}
\usepackage{amsmath}
\usepackage{svg}
\usepackage{amsfonts}
\usepackage{amssymb}
\usepackage{xcolor}
\usepackage{tcolorbox}
\usepackage{enumitem}

\usepackage{csquotes}

\usepackage{todonotes}

\definecolor{anne}{rgb}{0,0.5,0.9}
\definecolor{jakob}{rgb}{0.998,0.722,0.635}
\definecolor{instr}{rgb}{0.8,0.67,18}
\definecolor{heike}{rgb}{0,0.9,0.2}

\usepackage{soul,xcolor}
\setstcolor{red}

\newcommand{\setfootnotemark}{%
  \refstepcounter{footnote}%
  \footnotemark[\value{footnote}]}

\title{Texts or Images? A Fine-grained Analysis on the Effectiveness of \\ Input Representations and Models for Table Question Answering}

\author{Wei Zhou$^{1,3}$ \hspace{5mm}
  Mohsen Mesgar$^1$ \hspace{5mm}
  Heike Adel$^{2}$\hspace{5mm} 
  Annemarie Friedrich$^3$
    \\
  $^1$Bosch Center for Artificial Intelligence, Renningen, Germany \\ 
      $^2$Hochschule der Medien, Stuttgart, Germany\hspace{2.0mm} $^3$University of Augsburg, Germany \hspace{5mm} \\
\texttt{\{wei.zhou|mohsen.mesgar\}@de.bosch.com}\\ 
  \texttt{annemarie.friedrich@uni-a.de} \hspace{5mm} \texttt{adel-vu@hdm-stuttgart.de}}

\begin{document}
\maketitle

\begin{abstract}
In table question answering (TQA), tables are encoded as either texts or images.
Prior work suggests that passing images of tables to multi-modal large language models (MLLMs) performs comparably to or even better than using textual input with large language models (LLMs).
However, the lack of controlled setups limits fine-grained distinctions between these approaches.
In this paper, we conduct 
the first controlled study on the effectiveness of several combinations of table representations and models from two
perspectives: question complexity and table size.
We build a new benchmark based on existing TQA datasets.
In a systematic analysis of seven pairs of MLLMs and LLMs, we find that the best combination of table representation and model varies across setups.
We propose \textsc{fres}, a method selecting table representations dynamically, and observe a 10\% average performance improvement compared to using both representations indiscriminately.
\end{abstract}

\section{Introduction}
\label{sec:intro}

Table is a common data format in many downstream applications \citep{pasupat-liang-2015-compositional, parikh-etal-2020-totto} and domains \citep{chen-etal-2021-finqa, lu-etal-2023-scitab}. 
To process table data, current approaches either pass serialized table texts to large language models (LLMs) \cite{herzig-etal-2020-tapas, jiang-etal-2022-omnitab,liu2022tapex}, or input table images to multi-modal large language models (MLLMs) \cite{zheng-etal-2024-multimodal,alonso-etal-2024-pixt3}. 


\citet{deng-etal-2024-tables} show that these two approaches yield comparable performance on table-related tasks, such as table question answering (TQA), \textcolor{black}{when evaluated using GPT-4 \cite{achiam2023gpt} and $\text{Gemini}_{\text{Pro}}$ \cite{team2023gemini}. Sometimes, passing table images to MLLMs even outperforms inputting table texts to LLMs.}
However, the absence of controlled setups in prior work, e.g., the investigation of performance in relation to table size, hinders a deeper understanding of their strengths and weaknesses. 
\textcolor{black}{Moreover, existing investigations focus exclusively on closed-source models, raising the question of whether similar observations can be made using open-weights models of varying sizes.}

\label{sec:results}
\begin{figure}
    \centering
\includegraphics[width=0.9\linewidth]{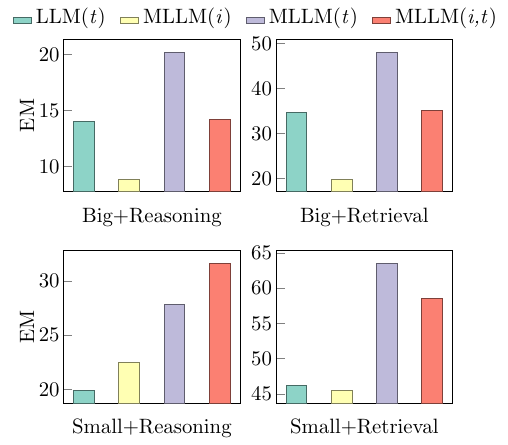}
    \caption{\textbf{Varying exact match (EM) for models and table representations under different settings} (\textit{i} and \textit{t} stand for image and text representations of tables). We categorize our investigation into four settings based on table size (small or big) and question complexity (retrieval or reasoning).}
    \label{fig:analysis_results}
\end{figure}
To study the effectiveness of different combinations of table representations (image vs.\ text vs.\ a combination of both) and models (LLMs vs.\ MLLMs)  under controlled circumstances, we build a \textbf{new benchmark} with four controlled settings sourced from six widely used TQA datasets. 
Our evaluation benchmark consists of 1600 instances, with tables in both image and text formats. 
We annotate instances with question complexity (retrieval vs.\ reasoning) and table size (small vs.\ big). 
In our \textbf{experimental study,}
we carefully evaluate seven open-weights MLLMs with their corresponding pre-trained LLM decoders in a zero-shot way. 
\textcolor{black}{Our results indicate that model size indeed plays a role in determining the most effective model-representation combinations: using the large Qwen-2-72B model, we find that passing table images consistently outperforms passing table texts, aligning with observations reported by \citet{deng-etal-2024-tables}. However, for small models (with $\le$12B parameters), the optimal combination of model and table representation varies across different settings, as illustrated in Figure \ref{fig:analysis_results}.}
For instance, MLLMs operating on both table text and image is the most effective approach when tables are small and questions require reasoning to be solved.
\textcolor{black}{Key findings are presented at the end of Section \ref{key_messages}.}
Finally, we propose \textbf{f}eature-based table \textbf{re}presentation \textbf{s}election, \textbf{\textsc{Fres}}, a method determining the best table representation \textcolor{black}{for small MLLMs} based on question complexity and table size.
By applying \textsc{Fres}, we observe an average of 10\% exact match gain 
compared to baseline approaches.
Code and data are available.\footnote{\url{https://github.com/boschresearch/FRES}}

\section{Benchmark Creation}
\label{sec:benchmark}
We build a new evaluation benchmark consisting of 1600 instances from six common TQA datasets. 
The benchmark features two dimensions and provides four controlled settings, with each setting comprising 400 instances. 
Table \ref{tab:dataset_features} lists the number of instances collected 
for each investigated setting. 

\begin{table}[!t] 
    \small
    \setlength{\tabcolsep}{3pt}
    \centering    
    \begin{tabular}{@{}llc@{}}
    \toprule
    \textbf{Settings} & \textbf{Source Dataset (\#Instance)} & \textbf{\#Total}\\
    \midrule
    Retrieval \\ 
    \hspace{2mm} Small&  WTQ (100), TabFact (100), HiTab (200)  & 400 \\
    \hspace{2mm} Big  & WTQ (100), TabFact (100),  HiTab (200) &400\\
    \midrule
    Reasoning \\
    \hspace{2mm} Small & \begin{tabular}{l}
            WTQ (50), TabFact (50),  HiTab (200)\\ TabMWP (50), CRT (50)
    \end{tabular}& 400 \\
     \hspace{2mm} Big &  \begin{tabular}{l}
          WTQ (50),TabFact (50), HiTab (200)\\TempTabTQA (50),CRT (50)
     \end{tabular}  & 400 \\ 
    \bottomrule
    \end{tabular}
       \caption{Our controlled settings are based on two comparison dimensions: question complexity (retrieval vs. reasoning) and table size (small vs big). The middle column shows the source datasets and the number of their instances in our benchmark.}
       \label{tab:dataset_features} 
\end{table}

\subsection{Source Datasets} 
We build up our benchmark from the development sets of six commonly-used single-table TQA datasets. 
We use table images and texts already provided in MMTab \citep{zheng-etal-2024-multimodal} for four out of six datasets:
WTQ \cite{pasupat-liang-2015-compositional}, TabFact \cite{2019TabFactA}, TabMWP \cite{Lu2022DynamicPL}, and HiTab \cite{cheng-etal-2022-hitab}.
These datasets feature numerical reasoning questions.
To broaden the range of reasoning types, we also include TempTabTQA \cite{gupta-etal-2023-temptabqa} for temporal reasoning and the test-only dataset CRT \cite{zhang-etal-2023-crt} for commonsense reasoning. 
\textcolor{black}{Note that TabMWP exclusively features small tables; we use it to collect instances with small tables. 
Similarly, TempTabTQA predominantly features large tables. We therefore use it to select instances with big tables.}
In TabFact, each instance comprises a table and a statement about the table. The goal is to answer whether the statement is True or False.
We convert it into TQA format by decomposing statements into question-answer pairs (\ref{convert_tabfact}). 
We generate table images for TempTabTQA and CRT from table texts, using four templates and various color encodings with HTML rendering tools (\ref{template_and_layout}).
Dataset statistics are provided in \ref{dataset_analysis}. 

\subsection{Dimensions}
We categorize the instances in our benchmark by \textbf{question complexity} and \textbf{table size}.
These two dimensions 
offer orthogonal perspectives, i.e., from a task-instruction level and from an input-data level.\footnote{Although table structure (flat vs.\ hierarchical) could also be an interesting dimension, we do not condition for it because 
it correlates with table sizes: in our initial exploration, we find hierarchical tables are generally bigger than flat tables in existing table datasets. }
In terms of question complexity, questions that involve 
identifying a verbatim answer in 
a table are referred to as \textbf{retrieval questions}, while those requiring additional inferences, such as numerical, temporal, or commonsense reasoning, are called \textbf{reasoning questions}.
In terms of table size, we distinguish between \textbf{big} and \textbf{small} tables, motivated by the fact that
MLLMs have been found to struggle with high-resolution images in visual benchmarks \citep{Li2023MonkeyIR}.

\begin{table}[t] 
    \small
    \centering    
     \setlength{\tabcolsep}{2pt}
    \begin{tabular}{*{3}{c}}
    \toprule
    \textbf{MLLM} & \textbf{LLM} & \textbf{Size}\\
    \midrule
   \makecell{Qwen-2-VL\\ \cite{Wang2024Qwen2VLEV}}& \makecell{Qwen-2\\ \cite{Yang2024Qwen2TR}}& 7B / 72B\\
   \midrule
   Pixtral \cite{Agrawal2024Pixtral1}&Mistral-nemo\setfootnotemark\label{first} &12B\\ \midrule
   \makecell{Phi-3.5-vision-instruct\\ \cite{Abdin2024Phi3TR}} & Phi-3.5-mini & 4B \\
   \midrule
   \makecell{LlaVa-Next\\ \cite{Li2024LLaVANeXTInterleaveTM}} & \makecell{Mistral\\ \cite{Jiang2023Mistral7}}&7B \\
   \midrule
   \makecell{GLM-4v \cite{Zeng2024ChatGLMAF}} & GLM-4 & 9B  \\
   \midrule
InternVL2\setfootnotemark\label{second}&\makecell{Internlm2\_5-7B-chat\\ \cite{Cai2024InternLM2TR}} &8B\\
    \bottomrule
    \end{tabular}
    \caption{Selected MLLMs and LLMs. We omit citations of LLMs if they are the same as corresponding MLLMs'. \label{tab:models}} 
\afterpage{\footnotetext[\getrefnumber{first}]{\url{https://mistral.ai/news/mistral-nemo/}}
\footnotetext[\getrefnumber{second}]{\url{https://internvl.github.io/blog/2024-07-02-InternVL-2.0/}}}
\end{table}

\subsection{Data Collection Process}
\label{collect_process}
Our goal is to collect the same number of instances for each setting from the six source datasets. In total, we collect 1600 instances, with 400 for each setting. 
Statistics for each setting are shown in  \ref{dataset_analysis}.
First, we \textbf{distinguish between retrieval and reasoning questions}. 
Three out of six datasets (TabMWP, TempTabTQA, and CRT) exclusively feature reasoning questions.
For WTQ, we differentiate retrieval questions from reasoning questions using the classification model from \citet{zhou-etal-2024-freb} and use Qwen-2 72B \cite{Yang2024Qwen2TR} as the question classifier, which
achieves an accuracy of 93\% on a set of annotated instances from \citet{zhou-etal-2024-freb}. Details are available in \ref{question_type_classification}.
For TabFact and HiTab, we leverage the datasets' annotations on question types.

Next, we \textbf{distinguish between big and small tables.}
Thresholds determining size groups are calculated as the averages of number of pixels and number of table text tokens in MMTAB \cite{zheng-etal-2024-multimodal}.
We select MMTAB for threshold calculation as it features a vast collection of more than 120k tables originally from web pages and reports, targeting human readers.
Size distribution of different datasets in MMTab can be found in \ref{table_size_threshold}.
Tables with both pixel numbers and token numbers smaller than the average (2e6 and 288) are classified as small. 
Tables with both values larger than averages are categorized as big. 
The rest of the tables ($\sim$22\%) is not included in the benchmark.
This ensures each size group features big/small tables in both image and text representations.

\textcolor{black}{After distinguishing question complexity and table size, we rank instances with the same question type by their table sizes, and choose the top $k$ (shown in Table \ref{tab:dataset_features}) instances from each source dataset in ascending order to collect data with small tables and descending order for big tables. 
This way, we ensure that the distribution of data featuring big/small tables is distinctive.
We also keep a balance distribution of different table structures (800 instances with flat/hierarchical table).}

\section{Experiments}
We aim to compare different combinations of table representations and model types.
In this section, we present selected models and evaluation metrics, as well as discussing the results.

\subsection{Models and Evaluation}
 We select \textbf{seven open-weight} MLLMs and their corresponding pre-trained LLM decoders (shown in Table \ref{tab:models}) for reproducibility and generalizability.
The selected MLLMs contain a significantly smaller visual encoder compared to the LLM decoder.
This ensures MLLMs and LLMs share the same model structure with only minimal differences in parameters.

To mitigate pre-training data contamination, i.e., ensure that the curated evaluation data has not been 
greatly exposed to a model during pre-training, 
we only select a pair of MLLM and LLM if both feature a low accuracy  ($\le$ 20\%) in all four controlled settings when masking out questions, and when masking out tables. 
Model performances with questions or tables masked can be found in \ref{model_selection}.

In total, we select \textbf{six} pairs of \textbf{small} models (MLLMs and LLMs), and \textbf{one} pair of \textbf{big} models \textcolor{black}{(due to a limited number of large MLLMs satisfying the above model selection conditions).}
We use the same table layouts and prompt templates (\ref{prompts}) as previous work \cite{deng-etal-2024-tables, zheng-etal-2024-multimodal}.
During \textbf{evaluation}, we further exclude all instances that any of the models predicts correctly when only seeing the table or question. 
In total, 84\% of the evaluation data is preserved (at least 80\% per setting), ensuring sufficient amount of data available for evaluation.
We use exact match accuracy (EM) between reference answers and predicted answers as the evaluation metric and the 
Wilcoxon signed-rank test \cite{Wilcoxon1945IndividualCB} for testing statistical significance. 

\subsection{Results}
\label{main_results}
We present the averaged results across all six small models in Figure \ref{fig:analysis_results}. Results for individual models are shown in \ref{mllm_performance}. 
\textcolor{black}{Table \ref{tab:qwen72B_results} shows the results for Qwen-2-72B and Qwen-2-VL-72B.}

\textcolor{black}{Previous work suggests that passing table images to MLLMs can result in comparable or even better performances than inputting table texts to LLMs \citep{deng-etal-2024-tables}. We find that the observation is valid only with large models: in Table \ref{tab:qwen72B_results}, we observe that MLLM(\textit{i}) consistently leads to better performance than LLM(\textit{t}), regardless of question complexity and table size.}

\begin{figure}
    \centering
    \includegraphics[width=0.9\linewidth]{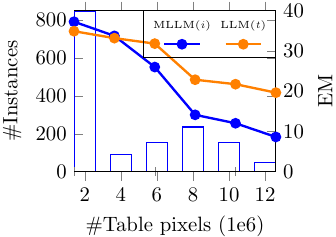}
    \caption{\textcolor{black}{Evaluation of table size robustness. The bar plot shows the number of instances sampled for each bin, and the line plots show the performance of different approaches against varying table sizes.}}
    \label{fig:res_dist}
\end{figure}

However, \textbf{the performance of these two approaches differs a lot under different settings when evaluating with small-sized models}: 
LLMs using table text (LLM(\textit{t})) outperform MLLMs using table images (MLLM(\textit{i})) when tables are large ($p<.05$).
\textcolor{black}{We believe this is because MLLMs struggle with processing large images \cite{Li2023MonkeyIR}. To prove that, we categorize instances into six distinctive bins according to their table sizes, and plot the EM of different approaches using grouped instances. This is shown in Figure \ref{fig:res_dist}. We find that table image representations are less robust compared to text representations with regard to table size. A similar figure plotted against table token numbers is shown in Figure \ref{fig:length_dist}, conveying the same message.}

For small tables, question complexity matters: MLLMs excel at reasoning questions using table images, while we do not find statistically significant differences for retrieval questions ($p=.67$). \textcolor{black}{This provides evidence for the hypothesis proposed in \citet{deng-etal-2024-tables}: ``representing tables as images can help LLMs in complex reasoning''. We suspect MLLMs' reasoning advantages come from training on massive image reasoning data. }

\begin{table}[!t]
    \small
    \centering  
     \setlength{\tabcolsep}{1pt}
    \begin{tabular}{lcccc}
    \toprule
    \textbf{Model} &\textbf{B/Reason} &  \textbf{B/Retriev}e&  \textbf{S/Reason}&  \textbf{S/Retrieve} \\\midrule
   LLM (\textit{t})& 29.2&69.5 &36.9&81.3  \\
   MLLM (\textit{i})& 46.5& 75.4 &67.1&87.7\\
    MLLM (\textit{t}) & 46.8 & \textbf{78.9}&60.7&88.1\\
    MLLM (\textit{i,t}) & \textbf{51.5} & 78.2&\textbf{70.0}&\textbf{89.4}\\
    \bottomrule
    \end{tabular}
       \caption{Exact Match of Qwen-2-72B on four different settings. \textit{i}, \textit{t} refer to tables as images and texts, respectively. B and S stand for Big and Small in terms of table size, respectively. \label{tab:qwen72B_results}}
\end{table}

\textbf{When comparing general capabilities of MLLMs and LLMs in solving TQA with textual inputs}, i.e., LLM(\textit{t}) versus MLLM(\textit{t}), we find that MLLMs outperform LLMs. 
It appears that continuing to pre-train and fine-tune LLMs with a small vision encoder using multi-modal data further enhances models' abilities in solving TQA.

\textbf{By comparing different table representations with MLLMs}, we observe that text-based approaches excel with large tables ($p<.05$). 
However with small tables, combining both table representations is the most effective choice for reasoning questions, while text alone suffices for retrieval questions ($p<.05$).
We suspect that passing both table representations better triggers the reasoning abilities of MMLMs, thus leading to superior performance on reasoning questions. 
However, in terms of information retrieval, the input is best represented as text,
probably due to the task being mostly presented in textual formats during instruction-tuning \cite{Zhu2023LargeLM}. 


\begin{tcolorbox}[enhanced,attach boxed title to top center={yshift=-3mm,yshifttext=-1mm},
  colback=green!5!white,colframe=green!75!black,colbacktitle=green!50!black,
  title=Key Takeaways,fonttitle=\bfseries,
  boxed title style={size=small,colframe=green!30!black}, left=1mm]
  \begin{itemize}[leftmargin=*,itemsep=1mm] 
     \item  For the large 72B model, table image is a more effective representation than table text. Passing both representations to the MLM results in the best performance.
      \item For small models:
       \begin{itemize}[leftmargin=1em,itemsep=0pt,topsep=0pt]
           \item with big tables, using text representations with MLLMs leads to the best performance. 
           \item with small tables, if the question type is reasoning, providing MLLMs with both table representations results in optimal performance. For retrieval questions, supplying  table texts to MLLMs suffices.  
       \end{itemize}
  \end{itemize}
\end{tcolorbox}
\label{key_messages}

\section{FRES: Feature-Based Table Representation Selection}

\begin{table}[t]
    \small
    \centering    

    \begin{tabular}{lcccc}
    \toprule
    \textbf{Methods} & \textbf{WTQ} & \textbf{TabFact}& \textbf{HiTab}& \textbf{WiKiSQL} \\
    \midrule
    Pxl-\textit{t}&51.3 & 75.7 & 59.7 & 58.8 \\
    Pxl-\textit{i} & 42.2& 74.2& 40.8 &51.7\\
    Pxl-\textit{t,i} &52.5 & \textbf{75.9}& 62.2 &60.1\\
    Pxl-\textsc{fres} &\textbf{54.4}&75.4&\textbf{64.0}&\textbf{61.4} \\
    \midrule
    TL-\textit{t}&34.4 & \textbf{63.7} &44.3 &47.8 \\
    TL-\textit{i} &19.2&59.5&10.5&31.6 \\
   TL-\textit{t,i} & 17.2&60.9&16.3&29.5\\
    TL-\textsc{fres} & \textbf{38.8} & 63.1 &\textbf{49.7}&\textbf{48.4}\\
    \bottomrule
    \end{tabular}
       \caption{Exact Match of models on TQA datasets. \textit{t}, \textit{i} refer to tables as texts and as images, respectively.  
       Pxl and TL refer to Pixtral and TableLlaVA, respectively.\label{tab:qa_results}}
\end{table}

Small MLLMs require less memory and are faster during inference than big MLLMs. 
We use observations found before and propose 
\textsc{Fres}, a method selecting the best table representations when using small-sized models:  
big tables are passed as texts to MLLMs; 
small tables are processed as texts when questions are retrieval-based; otherwise, both representations  are passed to MLLMs.
We employ the same classifier used in Section \ref{collect_process} to determine question types.
We regard tables with either pixels or token numbers larger the values in \ref{collect_process} as big. Otherwise, they are classified as small.

\paragraph{Datasets.}
We evaluate our method with the \textbf{test sets} of three aforementioned TQA benchmarks: WTQ, TabFact (small test) and HiTab for their diverse table sizes and question types.
We also include one additional dataset, WikiSQL \citep{zhongSeq2SQL2017}, that has not been used in our analysis to test the generalizability of our findings.
Dataset statistics are presented in \ref{dataset_analysis}.

\paragraph{Baselines and Models.}
We test four approaches: passing tables as \textbf{texts}, as \textbf{images}, as \textbf{both} representations, and the ones decided by \textsc{Fres}.
We choose the best performing MLLM from our previous analysis: \textbf{Pixtral 12b} (see \ref{mllm_performance} for individual model's performance) as well as a fine-tuned MLLM for TQA: \textbf{TableLlaVA 7B} \cite{zheng-etal-2024-multimodal}. 

\paragraph{Results.} Table \ref{tab:qa_results} displays the Exact Match of various approaches. 
Our proposed method, which selects input table formats based on question complexity and table size, proves to be effective. 
It achieves an average 10\% of EM gain across two models and four datasets, compared to using both representations indiscriminately.
In addition, it improves efficiency by reducing input token numbers (\ref{efficiency}).
We do not observe big differences between the baselines and our method on TabFact.
This might stem from the dataset's simplicity: TabFact is a classification dataset with only two labels.
It is likely to that differences between each representation become small or even diminish.

\paragraph{Error Analysis.}
\textcolor{black}{We analyze \textsc{fres} in terms of its failure patterns. To do that, we sample 200 instances where wrong predictions were given by Pixtral-7B with \textsc{fres}, with each test dataset 50 error instances. To investigate the impacts of table representations in causing errors, we obtain predictions of the same instances using different representations other than the one selected by \textsc{fres}. We note down changes in prediction correctness. If all representations fail to result in a correct prediction, we regard the error as coming from the \textbf{limited capability of the model} itself. As \textsc{fres} do not decide for image-only representations, any correct predictions obtained by changing to image-only table representations are regarded as \textbf{exceptional cases}.   
If changing table representations leads to correct predictions and the question type is predicted correctly, we regard the error as coming from \textbf{limitation of table size thresholds}. Otherwise, we regard the error coming from \textbf{failures in the question classification model}.
To obtain correct question type labels, we manually annotate the 200 instances. 
We find 65\% of the errors fall into the category of limited model capability, while approximately 10\% exceptional cases in which our observations do not capture. Around 19\% of the errors are stemming from question classification errors,  and 7\% from limitations of table size thresholds.}

\section{Conclusions}
\label{sec:conclusions}
%
We presented
a new benchmark for exploring the impact of using different input formats (text vs.\ image) with LLMs vs.\ MLLMs conditioned on question complexity and table size. 
In a systematic evaluation of seven models, we find that each combination of table representation and model type performs differently under different settings.
We propose \textsc{fres}, a method integrating our findings to dynamically select the best table representation.
\textsc{fres} effectively improves TQA performance. 


\section*{Limitation}
This study primarily focuses on table question answering, motivated by the availability of extensive datasets. 
However, it could be pertinent to investigate whether our findings are applicable to other table-centric tasks, such as table summarization. 
Moreover, we only explore the application of our findings in a simple setting where MLLMs are prompted in a zero-shot manner.
Future work can develop more intricate methods leveraging both modalities with our findings. 
Lastly, we assume the presence of tables in both image and textual formats for our initial exploration. 
Yet, in practical applications, the conversion between these representations may result in the loss of information, e.g., OCR might not work well when converting an image of a big table image to corresponding textual representation \cite{zheng-etal-2024-multimodal}. 

\bibliography{custom}

\newpage
\appendix
\section{Appendix}

\begin{table*}[htb]
    \small
    \centering    
 
    \begin{tabular}{*{8}{c}}
    \toprule
    Dataset & Q Length & T Length & A Length&T Resolution & Q Complexity & T Structure&\#Samples\\\midrule
    TABMWP&25.1 & 21.7&1&2.4e4&numerical reasoning&flat&50  \\
    TempTabTQA &13.4 &461.9 &1.7&3.8e6&temporal reasoning &flat&50 \\
    CRT &24.2 &304.8&1.3&4.6e6&mixed reasoning&flat&100  \\
    WTQ &10.2 &676 &1.9&3e6& retrieval, reasoning&flat&300\\
    TabFact& 11.3&320 & 1.8&3.4e6& retrieval, reasoning&flat&300\\
    HiTab& 17.6& 425.8 &1.6&4.7e6& retrieval, reasoning&hierarchical&800 \\
    All &15.6 & 433.9&1.7&4e6 & retrieval, reasoning&flat, hierarchical&1600\\
    \bottomrule
    \end{tabular}
       \caption{General dataset features and statistics. Q, T, and A stands for question, table, and answer, respectively. Lengths and resolutions are averaged over collect samples. Q, T, and A lengths are calculated as the number of white-space separated tokens. \label{tab:general_statistics}}
\end{table*}

\begin{table*}[htb]
    \small
    \centering    
 
    \begin{tabular}{*{6}{c}}
    \toprule
    Setting & Q Length & T Length & A Length&T Resolution &\#Samples\\\midrule
    Retrieve-Small&13.3 & 105.9&1.5&6.8e5&400  \\
    Retrieve-Big &14.3 &844.5 &1.9&7.1e6&400 \\
    Reasoning-Small &18.0 &93.0&1.5&6.8e5&400  \\
    Reasoning-Big &16.7 &692.1 &1.9&7.6e6&400\\
    \bottomrule
    \end{tabular}
       \caption{Statistics per evaluation setting. Q, T, and A stands for question, table, and answer, respectively. Lengths and resolutions are averaged over collected samples. Q, T, and A lengths are calculated as the number of white-space separated tokens. \label{tab:setting_statistics}}
\end{table*}


\subsection{Converting TabFact to a TQA Setting}
\label{convert_tabfact}
Each instance in TabFact comprises a table, a statement about the table, and an answer if the statement is True or False.
We select only the true statements to ensure that the answers contained in the statements are correct. 
Subsequently, we use GPT-4 to decompose these statements into question-answer pairs. 
We exclude instances whose answers are not found in the original statements. 
To test the validity of the created question answering dataset, we run GPT-4 three times and choose the most frequent answers as the predicted answers.
By comparing the predicted answers and the gold answers obtained from GPT-4 by statement decomposing, an accuracy of 90\% is reached.
In cases of errors, we find that either the evaluation is challenging (due to free-form text) or the original statements are inaccurate. 
This suggests the validity of the question answering dataset created from TabFact.

\subsection{Creating Table Image}
\begin{figure}
  \begin{subfigure}[t]{.4\columnwidth}
    \centering
    \includegraphics[width=\linewidth]{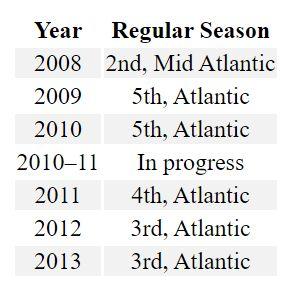}
    \caption{Table without cell border/DataFrame-like.}
  \end{subfigure}
  \hfill
  \begin{subfigure}[t]{.5\columnwidth}
    \centering
    \includegraphics[width=\linewidth]{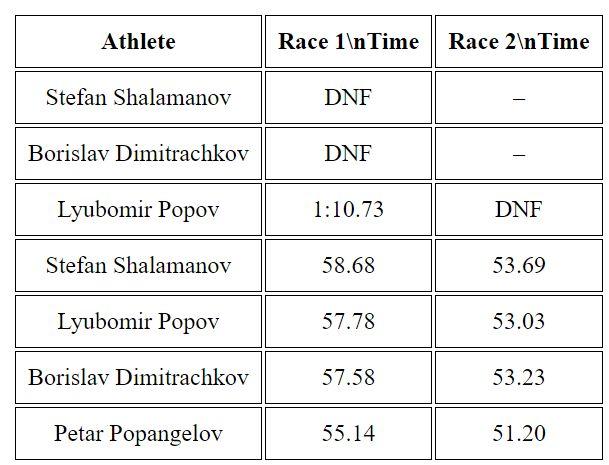}
    \caption{Table with cell boarders.}
  \end{subfigure}

  \medskip

  \begin{subfigure}[t]{.35\columnwidth}
    \centering
    \includegraphics[width=\linewidth]{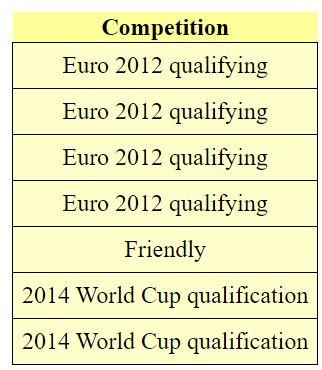}
    \caption{Table with partial borders.}
  \end{subfigure}
  \hfill
  \begin{subfigure}[t]{.58\columnwidth}
    \centering
    \includegraphics[width=\linewidth]{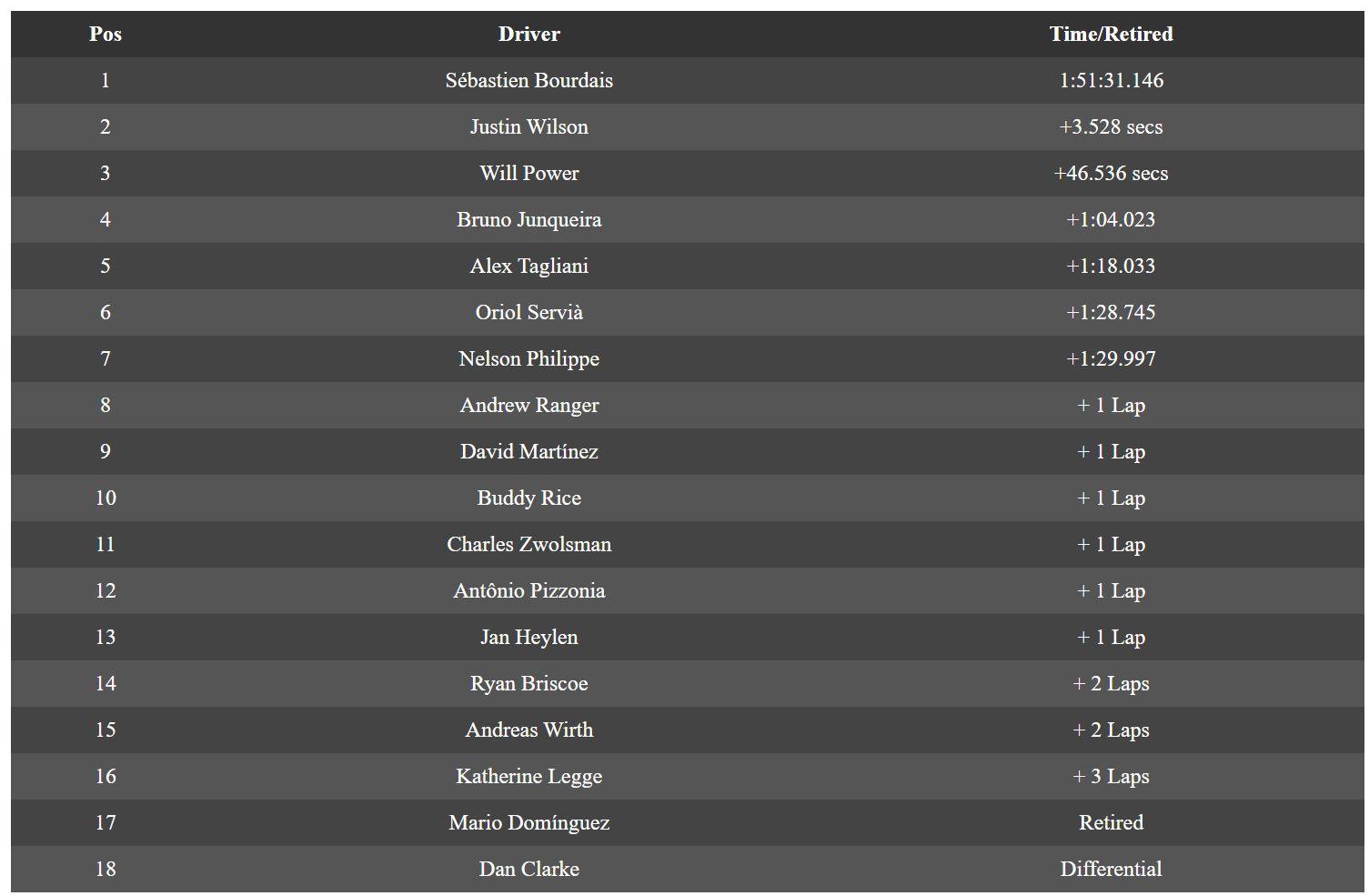}
    \caption{Table with wide cell space.}
  \end{subfigure}
\caption{Different table templates.}
\label{fig:table_templates}
\end{figure}

 We use four templates to create table images given table text. The templates are shown in Figure \ref{fig:table_templates}. 
To create a table image, we first represent tables as Pandas DataFrame object, then we parse a DataFrame into HTML with the function \texttt{df.to\_html}. We convert HTML to image with Python package \texttt{Html2Image}.\footnote{https://pypi.org/project/html2image/} Lastly, we crop images to eliminate redundant white space. 

\label{template_and_layout}

\subsection{Dataset Collection and Statistics}
\label{dataset_analysis}

\paragraph{Evaluation Data Statistics.}
To create our evaluation data, we have to distinguish retrieval and reasoning questions, as well as small and big tables. We present an analysis of our collected evaluation data in two aspects: \textbf{statistics per source dataset}, and \textbf{statistics per evaluation setting}. These are shown in Table \ref{tab:general_statistics} and \ref{tab:setting_statistics}, respectively.
\paragraph{Test Data Statistics.} We present statistics for four test datasets in Table \ref{tab:general_statistics_test}. We show averages of cell number, table tokens, and resolution, together with the percentage of small tables, reasoning questions, and instances featuring both small tables and reasoning questions. As no annotation of question complexity exists, the statistics are calculated based on the question classifier explained in \ref{question_type_classification}. The categorization of small and large tables is based on heuristics calculated from MMTab (\ref{table_size_threshold}). 

\paragraph{Dataset Licenses}
We build our evaluation dataset based on a subset of MMTab \cite{zheng-etal-2024-multimodal}, CRT \cite{zhang-etal-2023-crt} and TempTabTQA \cite{gupta-etal-2023-temptabqa}. 
The three datasets are publicly available under the licenses of \textsc{Apache-2.0}\footnote{\url{https://opensource.org/license/apache-2-0}}, \textsc{MIT}\footnote{\url{https://opensource.org/license/mit/}} and \textsc{CC-BY-4.0}\footnote{\url{https://creativecommons.org/licenses/by/4.0/legalcode}}, respectively.
In terms of the test datasets: WTQ \cite{pasupat-liang-2015-compositional}, TabFact \cite{2019TabFactA}, HiTab \cite{cheng-etal-2022-hitab} and WikiSQL \cite{zhongSeq2SQL2017}, they are under the license of \textsc{CC-BY-SA-4.0}\footnote{\url{https://creativecommons.org/licenses/by-sa/4.0/}}, \textsc{MIT}, \textsc{BSD-3 Clause}\footnote{\url{https://opensource.org/license/bsd-3-clause}} and \textsc{C-UDA}\footnote{\url{https://github.com/microsoft/HiTab?tab=License-1-ov-file}} respectively.
\subsection{Question Type Classification}
\label{question_type_classification}
To distinguish retrieval and reasoning questions in \textbf{TabFact}, we use the simple and complex splits to collect instances, respectively. 
For \textbf{HiTab}, we utilize the dataset's annotations to differentiate between retrieval and reasoning questions; specifically, questions categorized with an aggregation type of \textit{None} are considered retrieval questions, while all others are deemed reasoning questions.
For \textbf{WTQ}, we utilize the question type classifier proposed in \citet{zhou-etal-2024-freb}.
More specifically, a rule-based method is applied first: if an answer is not in a table, a question is classified as a reasoning question. If a question contains comparative terms (we detect it using NLTK), the question is classified as a reasoning question.
Next, an LLM takes in a question and a table and returns a predicted question type. 
We replace the LlaMA 2 (13b) used in the original paper with Qwen 2(72b) for its better general capabilities, but keep the prompt the same. 
We examine the validity of the question classifier by testing it on 200 instances from WTQ, annotated with gold question type \citep{zhou-etal-2024-freb}. 
This results in an accuracy of 93\%.

\subsection{Table Size threshold}
\label{table_size_threshold}
\begin{figure}
    \centering
    \includegraphics[width=0.9\linewidth]{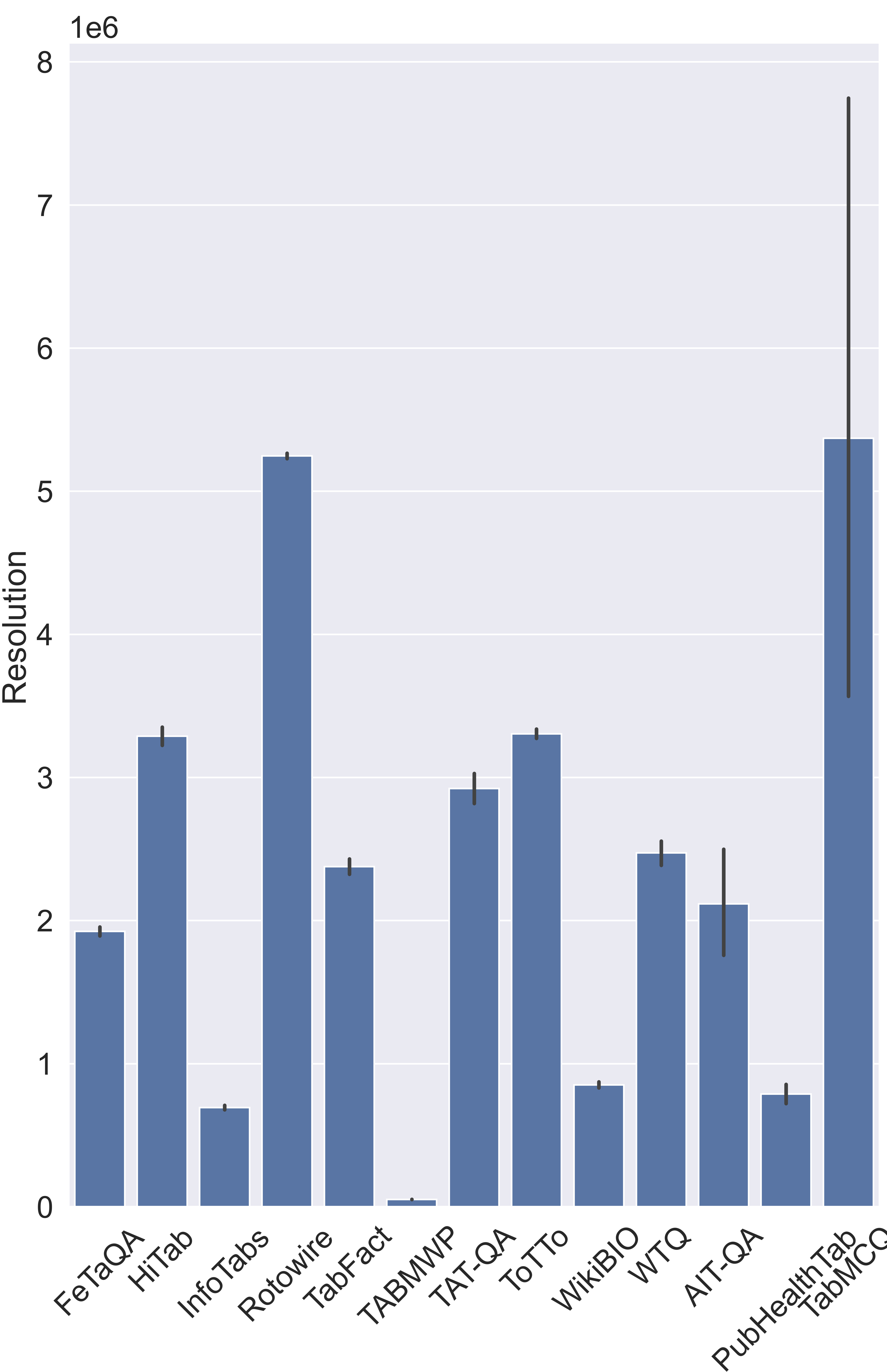}
    \caption{Resolution distribution of MMTab.}
    \label{fig:mm_dist}
\end{figure}

We use the entire MMTab dataset \cite{zheng-etal-2024-multimodal}, consisting of more than 120 thousand table images to calculate the averaged table image resolution. 
This results in a value of 2e6 (width*height). 
The resolution distribution is shown in Figure \ref{fig:mm_dist}. 
We average all token numbers in the table texts presented in MMTab. This results in an average of 288 tokens and 120 cells.

\begin{figure}[!h]
    \centering
    \includegraphics[width=0.98\linewidth]{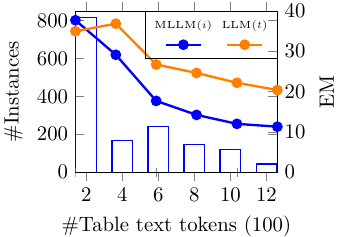}
    \caption{Evaluation of table size robustness. The bar plot shows the number of instances sampled for each bin and the line plots show the performance of different approaches against varying table sizes.}
    \label{fig:length_dist}
\end{figure}

\subsection{Model Selection}

\label{model_selection}
Table \ref{tab:mask_out_results} shows the number of instances that models still predict correctly when masking out questions or tables under different settings. 
We eliminate models when either MLLMs or corresponding LLMs can predict 20\% of the data correctly in any setting. 
This means we do not consider models that can predict more than 400*0.2=80 instances. All six examined models passed the test. 
\begin{table*}[!h]
\centering
\resizebox{\textwidth}{!}{%
  \begin{tabular}{lcccccccc}
    \toprule
    \multirow{2}*{\textbf{Models}} & \multicolumn{4}{c}{\textbf{No Question}} & \multicolumn{4}{c}{\textbf{No Table}}  \\
      \cmidrule(lr){2-5}
      \cmidrule(lr){6-9}
      & {Big/Reasoning} & {Big/Retrieve} & {Small/Reasoning} & {Small/Retrieve} & {Big/Reasoning} & {Big/Retrieve} & {Small/Reasoning} & {Small/Retrieve} \\
      \cmidrule(lr){1-5} \cmidrule(lr){6-9}
    Mistral-7b & 9&6&16&8&16&34&4&18 \\
    LlaVA-Next &0&0&3&0&21&49&4&24 \\
    \midrule
    Mistral-Nemo-12b &2&1&9&2&19&62&6&27 \\
    Pixtral-12b &0&0&1&0&37&57&3&24 \\
    \midrule
     Qwen2-7b &10&9&32&2&25&53&4&19 \\
    Qwen2-VL-7b &2&2&3&0&29&61&4&30 \\
    \midrule
    Phi-3.5-mini&4&0&4&2&4&11&0&6  \\ 
     Phi-3.5-vision&1&2&1&0&14&37&2&14\\
      \midrule
     GLM-4-9b &0&1&7&1&3&31&0&10 \\
    GLM-4v-9b &2&3&13&1&28&28&8&26 \\
    \midrule
       Internlm2\_5-7b&7&4&14&3&10&34&0&8 \\
    InternVL2-8b & 1&2&3&0&24&66&3&26\\
    \bottomrule
  \end{tabular}}
\caption{Number of instances that models still correctly predict when making out questions or tables.}
\label{tab:mask_out_results}
\end{table*}

\subsection{Prompts}
\label{prompts}
\begin{figure*}
    \centering
\includegraphics[width=0.98\linewidth]{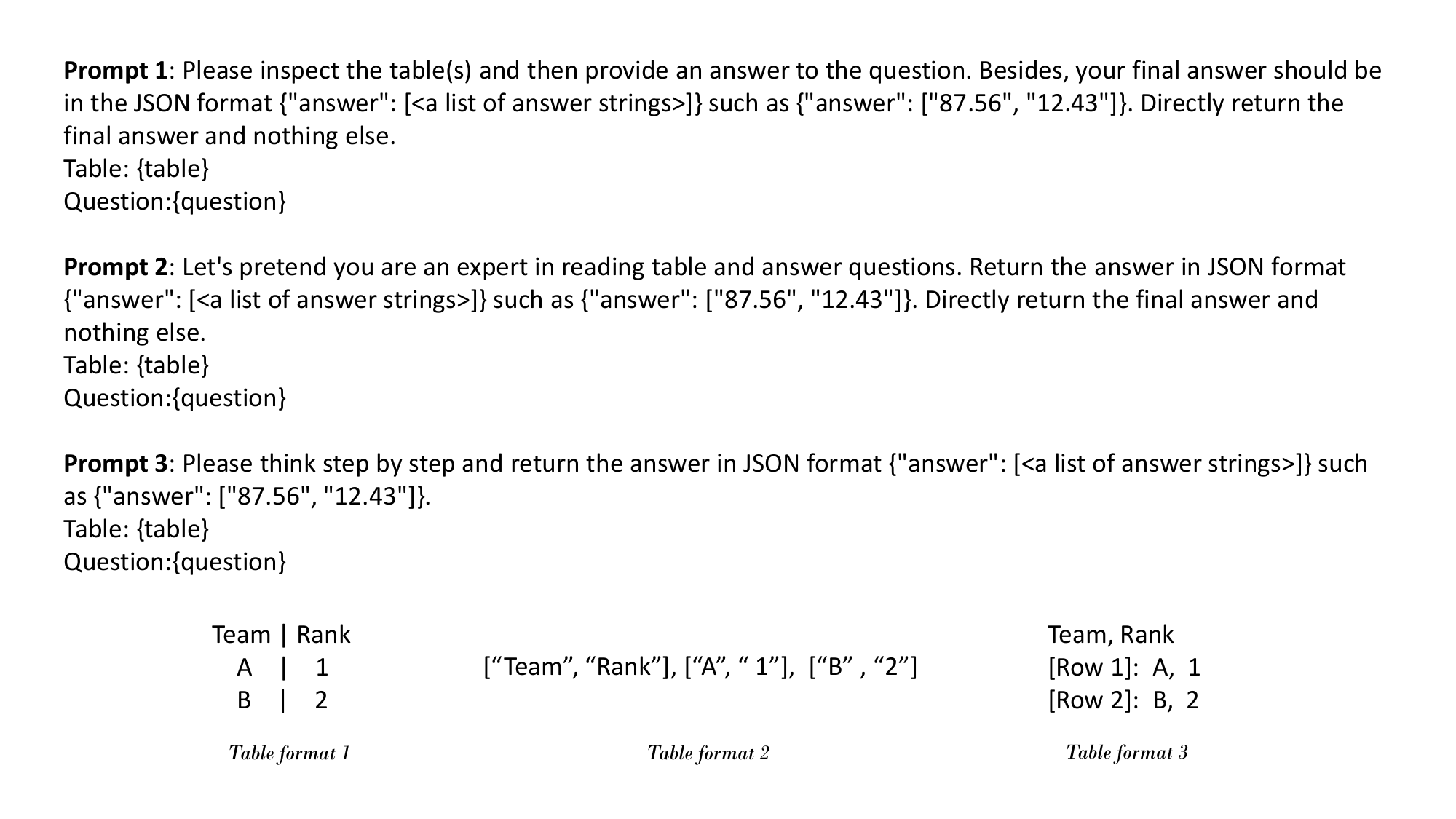}
    \caption{Prompts and table formats.}
    \label{fig:prompts}
\end{figure*}

We present prompts and table formats used in this study in Figure \ref{fig:prompts}.   

\begin{table*}[htb]
    \small
    \centering    
 
    \begin{tabular}{*{7}{c}}
    \toprule
    Dataset & \#cell & \#table-text & \#table-img& \%small table &\%reasoning question& \%small\_reasoning\\\midrule
    WTQ&165&429&2.4e6 & 57.7&71.4&41.6  \\
    TabFact &94 & 258&1.8e6&69.3&51.2&35.8 \\
    HiTab & 180&360 & 3.5e6&54.5&46.0&26.0 \\
    WikiSQL& 95& 227& 1.7e6& 71.3&29.8&22.0 \\
    \bottomrule
    \end{tabular}
       \caption{Statistics of test datasets. We show the average number of cells, white-space-separated table tokens, and image resolutions (height*width). In addition. Percentages of small tables, reasoning questions as well as a combination of both are also presented.\label{tab:general_statistics_test}}
\end{table*}

\subsection{MLLM Result}
\label{mllm_performance}

Table \ref{tab:individual_performance} shows the performances of individual MLLMs on the evaluation set. 
We find that Pixtral 12B performs the best among all evaluated small models. 

\begin{table*}
    \small
    \centering    
    \begin{tabular}{*{6}{c}}
    \toprule
    Model&Setting &Big/Reasoning &  Big/Retrieval&  Small/Reasoning&  Small/Retrieval \\\midrule
   \multirow{4}*{Qwen2 7b}& LLM (\textit{t})&15.6 &41.3 &20.6&59.7  \\
   & MLLM (\textit{i})& 13.1& 25.2 &26.8&72.2\\
  &  MLLM (\textit{t}) &21.6  &54.5 &35.3&73.6\\
   & MLLM (\textit{i,t}) &9.1 &20.7&36.0&72.0\\
   \midrule
   \multirow{4}*{Pixtral 12b}& LLM (\textit{t})& 19.4&58.7 &29.0& 73.3 \\
   & MLLM (\textit{i})&17.3 & 36.9 &41.7&71.6\\
  &  MLLM (\textit{t}) &\textbf{30.3}  & 63.1&46.1&\textbf{75.6}\\
   & MLLM (\textit{i,t}) &29.8 &\textbf{65.7}&\textbf{48.5}&73.5\\
   \midrule
   \multirow{4}*{Phi-3.5 4b}& LLM (\textit{t})&12.9 &39.8 &20.9&55.5  \\
   & MLLM (\textit{i})& 11.2&35.6  &19.4&58.7\\
  &  MLLM \textit{t}) & 18.3 & 46.2&21.4&68.3\\
   & MLLM (\textit{i,t}) &16.6 &40.3&36.1&67.2\\
   \midrule
   \multirow{4}*{LlaVA 7b}& LLM (\textit{t})&12.9 &33.1 &11.7& 38.7 \\
   & MLLM (\textit{i})&1.3& 3.3 &4.2&14.7\\
  &  MLLM (\textit{t}) &14.6  &44.0 &15.4&59.0\\
   & MLLM (\textit{i,t}) &7.4 &24.0&16.6&45.6\\
   \midrule
 \multirow{4}*{GLM-4 9b}& LLM (\textit{t})& 11.6&18.3 &12.7& 24.3 \\
   & MLLM (\textit{i})&5.9& 12.0 &14.9&27.7\\
  &  MLLM (\textit{t}) & 16.7 & 25.8&13.7&36.9\\
   & MLLM (\textit{i,t}) &2.9 &9.3&15.5&27.4\\
   \midrule
 \multirow{4}*{Intern 8b}& LLM (\textit{t})&11.7 &17.4 &24.6& 25.7 \\
   & MLLM (\textit{i})&4.7 & 6.6 &28.0&43.1\\
  &  MLLM (\textit{t}) & 19.8 &55.0 &35.1&67.7\\
   & MLLM (\textit{i,t}) &19.1 &51.4&36.8&65.2\\
   \midrule
  \multirow{4}*{Average}& LLM (\textit{t})&14.0 &34.8 &19.9& 46.2 \\
   & MLLM (\textit{i})&8.9 &  19.9&22.5&45.5\\
  &  MLLM (\textit{t}) & 20.2 &48.1&27.8&63.5\\
   & MLLM (\textit{i,t}) &14.2 &35.2&31.6&58.5\\

    \bottomrule
    \end{tabular}
       \caption{Exact Match of different models on four different settings. \textit{i}, \textit{t} refer to tables as image and text representations.\label{tab:individual_performance}}
\end{table*}


\subsection{Efficiency Analysis of FRES}
\label{efficiency}
We compare \textsc{fres} with passing both table and image representations to MLLMs indiscriminately in terms of efficiency.
\textsc{fres} avoids passing table images for around 80\% of data in the four test sets.
This results in up to  66\% fewer input token numbers over the six investigated small-size models, compared to passing both representations.

\end{document}